\documentclass[lettersize,journal]{IEEEtran}

\usepackage{framed}
\usepackage{booktabs}
\usepackage{subfigure}
\usepackage{amsmath}
\usepackage{graphicx}
\usepackage[justification=centering]{caption}

\begin{document}

\title{Data Trajectory Alignment for LLM Domain Adaptation: A Two-Phase Synthesis Framework for Telecommunications Mathematics}

\author{
	Zhicheng Zhou,
	Jing Li \\
	Suming Qiu,
	Junjie Huang,
	Linyuan Qiu,
	Zhijie Sun \thanks{Corresponding author: Zhijie Sun. Email: sunzhijie3@huawei.com} \\
	Global Technical Service (GTS) \\
	Huawei Technologies Co., Ltd
}

\maketitle

\begin{abstract}
	General-purpose large language models (LLMs) are increasingly deployed in verticals such as telecommunications, where adaptation is hindered by scarce, low–information-density corpora and tight mobile/edge constraints. We propose \emph{Data Trajectory Alignment} (DTA), a two-phase, model-agnostic data curation framework that treats solution \emph{processes}—not only final answers—as first-class supervision. Phase~I (\emph{Initializing}) synthesizes diverse, high-coverage candidates using an ensemble of strong teachers. Phase~II (\emph{DTA}) rewrites teacher solutions to \emph{align} intermediate steps and presentation style with the target student’s inductive biases and then performs signal-aware exemplar selection via agreement checks and reflection-based judging. 
	
	Instantiated on \emph{telecommunications mathematics} (e.g., link budgets, SNR/AMC selection, and power-control feasibility), DTA yields state-of-the-art (SOTA) accuracy on \textsc{Telemath} without enabling explicit “thinking” modes: \textbf{72.45\% pass@1}, surpassing distilled-only training by \textbf{+17.65} points and outperforming a strong baseline (Qwen3-32B with thinking enabled) by \textbf{+2.94} points. Token-shift analyses indicate that DTA concentrates gains on logical–structural discourse markers rather than merely amplifying domain nouns, indicating improved reasoning scaffolding. 
	
	Under edge-like inference settings, DTA improves efficiency by reducing reliance on multi-sample voting and disabling expensive reasoning heuristics, cutting energy per output token by \textbf{$\sim$42\%} versus Qwen3-32B (thinking mode enabled) and end-to-end latency by \textbf{$\sim$60\%} versus Qwen3-32B (thinking mode disabled). These results demonstrate that aligning \emph{how} solutions are produced enables compact, high-yield supervision that is effective for both accuracy and efficiency, offering a practical recipe for domain adaptation in low-resource verticals beyond telecom.
\end{abstract}


\section{Introduction}
General-purpose large language models (LLMs) are rapidly being deployed across verticals such as law, healthcare, and telecommunications \cite{xu2025staf}. However, adapting generic models to telecom tasks remains difficult for two coupled reasons. First, domain supervision is \emph{scarce} and often \emph{low–information-density}: expert knowledge tends to appear as terse formulas, parameter tables, and semi-structured artifacts rather than richly annotated text \cite{chen-etal-2021-finqa}. Second, mobile/edge deployments impose tight budgets on compute, memory bandwidth, and energy, which disfavor inference-time “heavy” reasoning procedures \cite{sze2017efficientdnns, hennessy2011computer}. The central question is thus: \emph{How can we obtain high-yield supervision that enables accurate, efficient telecom reasoning when both labeled data and inference resources are constrained?}

A common response is to synthesize training data via \emph{distillation} from strong teacher models \cite{fang2025knowledge}. While distillation boosts data \emph{quantity} and sometimes quality, it leaves a critical gap: the \emph{trajectory}—the tone, organization, and granularity of intermediate steps—often reflects the teacher’s habits rather than the student’s inductive biases. This mismatch, which we term \emph{trajectory debt}, can cause brittle transfer and poor one-shot accuracy, even when final answers in the synthetic corpus are correct \cite{lai2024stepdpostepwisepreferenceoptimization}. In practice, debt manifests as over-verbose rationales, unstable step demarcation, or missing constraint checks—failure modes that are especially harmful for telecom mathematics (e.g., link budgets, AMC selection, power-control feasibility), where unit consistency and engineering constraints must be enforced.

\begin{figure*}[t]
	\centering
	\includegraphics[width=0.9\textwidth]{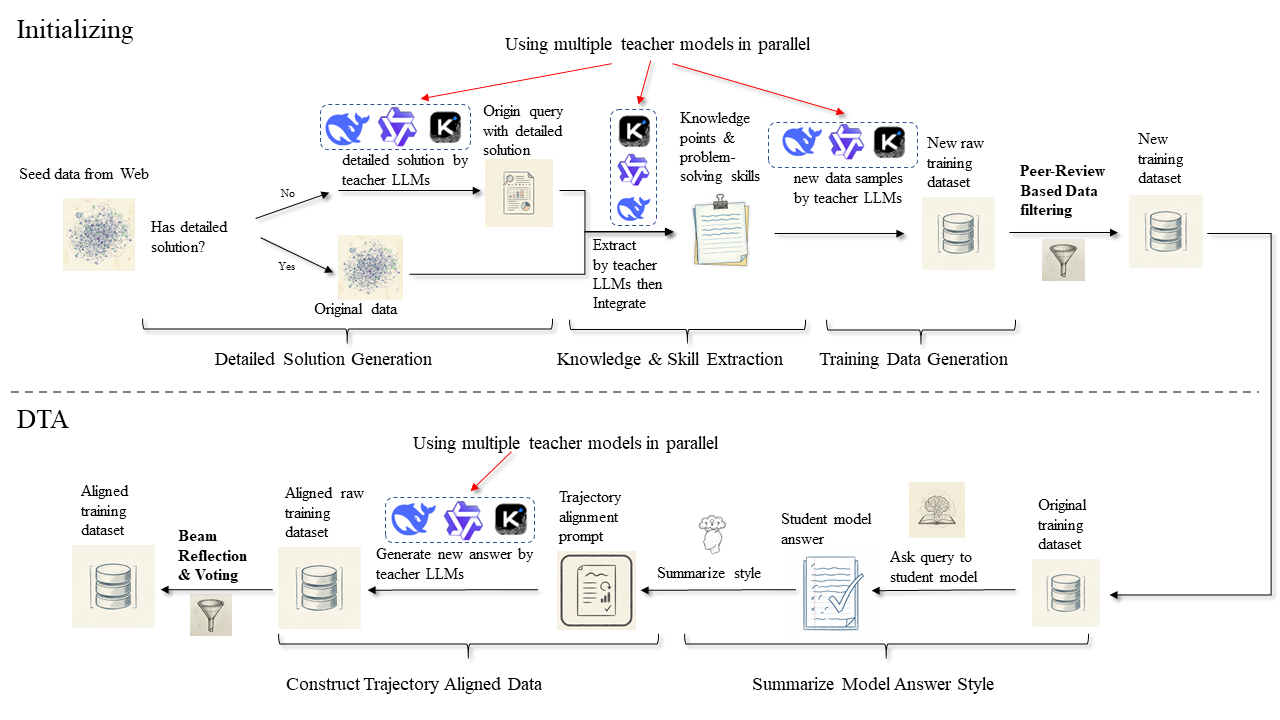}
	\caption{The working flow of two-phase framework: initializing and DTA}
	\label{flow}
\end{figure*}

We propose \emph{Data Trajectory Alignment} (DTA), a two-phase, model-agnostic data curation framework that treats solution \emph{processes}—not only final answers—as first-class supervision. Phase~I (\emph{Initializing}) uses an ensemble of strong teachers to synthesize diverse, high-coverage candidates. Phase~II (\emph{DTA}) explicitly \emph{aligns} the intermediate steps and presentation style to the target student, followed by signal-aware exemplar selection driven by agreement checks and reflection-based judging. The pipeline is largely automated (consistency/self-agreement filters) and plugs into standard supervised fine-tuning without requiring tool-use or retrieval. An overview is shown in Figure~\ref{flow}.

By aligning \emph{how} solutions are produced, DTA improves \emph{pass@1} without enabling explicit “thinking” modes, thereby reducing reliance on multi-sample voting and expensive reasoning heuristics. This yields concrete systems benefits under edge-like settings: lower latency, lower energy per token, and higher throughput—aligning the data curation objective with mobile/edge performance constraints rather than working against them.

Instantiated on \emph{telecommunications mathematics} and evaluated on \textsc{Telemath} \cite{colle2025telemathbenchmarklargelanguage}, DTA achieves state-of-the-art accuracy \emph{without} explicit thinking at inference: \textbf{72.45\% pass@1}, outperforming distilled-only training by \textbf{+17.65} points and a strong baseline (Qwen3-32B with thinking enabled) by \textbf{+2.94} points (Table~\ref{tab:leaderboard}). Under edge-like inference, DTA reduces energy per output token by \textbf{$\sim$42\%} versus the thinking-enabled baseline and cuts end-to-end latency by \textbf{$\sim$60\%} versus the non-thinking baseline (Table~\ref{mobile}). Token-shift analyses indicate that DTA concentrates gains on \emph{logical–structural} discourse markers rather than merely amplifying domain nouns, evidencing improved reasoning scaffolding that stabilizes unit checks and constraint propagation.

This paper makes the following contributions:
\begin{itemize}
	\item \textbf{Formulation of Data Trajectory Alignment.} We introduce a two-phase framework that aligns step-wise \emph{solution trajectories} (tone, structure, level of detail) to the target student before training, directly addressing trajectory debt.
	\item \textbf{Model-agnostic and automated curation.} We combine cross-model agreement filters and reflection-based judging into a largely automated pipeline that yields compact, high-yield supervision from limited sources.
	\item \textbf{Mobile/edge relevance.} We demonstrate that trajectory-aligned supervision improves pass@1 \emph{without} explicit thinking, cutting energy and latency under edge-like settings—key metrics for TMC scenarios.
	\item \textbf{State-of-the-art telecom results and analyses.} On \textsc{Telemath}, DTA surpasses strong distilled and thinking-enabled baselines; ablations on general math show DTA’s benefits extend beyond telecom. Analyses (token-shift, representation drift) explain \emph{why} alignment improves first-pass correctness and robustness.
\end{itemize}

In summary, DTA provides a practical recipe for \emph{process-aligned} supervision that simultaneously advances accuracy and inference efficiency in low-resource, constraint-heavy domains such as telecommunications. The rest of the paper details the framework (Sec.~\ref{flow}), situates it within prior art (Sec.~\ref{sec:related}), and reports comprehensive experiments, ablations, and mobile/edge measurements.

\section{Related Work}
\label{sec:related}
Domain adaptation aims to transfer general LLMs to vertical tasks under limited, often low–information-density corpora. Prior approaches can be grouped into: (i) \emph{data-centric} adaptation, including continued pretraining/tuning on in-domain text and in-domain instruction tuning, which improves lexical/semantic coverage but can overfit surface statistics when step-wise structure is sparse; (ii) \emph{parameter-efficient} adaptation (e.g., adapter/LoRA-style SFT) \cite{houlsby2019adapter, hu2021lora}, which lowers training cost but inherits the supervision’s style and rationale quality; and (iii) \emph{retrieval-augmented} adaptation that injects external knowledge at inference, helpful when knowledge is explicit but less effective for numerical derivations requiring stable intermediate steps \cite{lewis2020rag}. In financial QA, for example, \cite{chen-etal-2021-finqa} demonstrates that domain grounding alone is insufficient without faithful multi-step reasoning. Our \emph{Data Trajectory Alignment} (DTA) targets a complementary axis: instead of solely enlarging or specializing corpora, it \emph{repairs the process} by aligning intermediate steps, tone, and organizational style to the \emph{student’s} inductive biases, thereby reducing trajectory mismatch when explicit “reasoning modes” are disabled at inference.

Public, standardized datasets for telecommunications engineering remain scarce relative to law/biomedicine. Real-world telecom materials (e.g., link-budget worksheets, 3GPP-style normative text, PHY/MAC parameter tables) are rich in formulas and units but are \emph{low-density} as supervision: many tokens are descriptive scaffolding rather than derivational signal, and gold rationales are rarely annotated. The \textsc{Telemath} benchmark \cite{colle2025telemathbenchmarklargelanguage} explicitly couples numerical reasoning with domain constraints (spectral efficiency, coding gains, error probabilities, power budgets), surfacing failure modes where models produce correct-looking formulas but violate units or feasibility constraints. Our setting adopts \textsc{Telemath} as seed data and shows that DTA-curated supervision lifts both pass@1 and self-consistency without enabling chain-of-thought at test time. 

Data distillation leverages a strong teacher to synthesize instruction–solution pairs for a smaller or cheaper student. Representative pipelines include Self-Instruct and Alpaca-style instruction bootstrapping \cite{wang2023selfinstruct,alpaca2023}, Evol-Instruct and its code/math variants WizardLM/WizardCoder/WizardMath that increase complexity/diversity \cite{xu2023wizardlm,luo2023wizardcoder,luo2023wizardmath}, and instruction back-translation that infers latent instructions for unlabeled corpora \cite{li2023instructionbacktranslation}. These methods primarily expand \emph{quantity/coverage}; however, because teachers and students differ in narration habits and step granularity, naive distillation can create a \emph{trajectory debt}: the synthesized rationale’s tone, structure, and detail level diverge from the student’s preferences, limiting transfer even when final answers are correct. 

Our framework complements distillation by inserting two curation stages before SFT: (i) \emph{style induction} from the student’s own outputs to form a prescriptive style guide, and (ii) \emph{trajectory rewriting} of teacher solutions into that guide, followed by \emph{signal-aware selection}. Concretely, we (a) run cross-model \emph{peer review} to filter hallucinations, (b) compute a dual \emph{reflection} score combining student-informativeness (the reversed-IFD criterion from Selective Reflection-Tuning \cite{li-etal-2024-selective}) with a reward-judge aggregation over correctness, completeness, clarity, and conciseness, and (c) keep high-yield exemplars. Token-shift analyses further show that DTA increases \emph{logical–structural} discourse markers relative to distilled-only data \cite{lin2023unlockingspellbasellms}, indicating that it regularizes the scaffolding needed for unit-consistent, constraint-aware derivations in telecommunications mathematics. Empirically, DTA improves over distilled-only training and closes the gap to larger reasoning-enabled baselines on \textsc{Telemath}, while also transferring to general math benchmarks in a distillation-free ablation.

In summary, where prior work emphasizes enlarging in-domain data, our DTA focuses on \emph{process compatibility}: aligning the \emph{how} of solutions to the target student before training. This yields compact, high-signal supervision that is particularly effective for telecom tasks characterized by scarce labels, dense formulas, and strict engineering constraints.

\section{Method}
\subsection{Initializing: Original Training Data Generation}
\subsubsection{Detailed Solution Generation}
Given a seed data sample $(x_{seed}, y_{seed})$ of problem $x_{seed}$ and solution $y_{seed}$ from the web, to integrate the capabilities of multiple high-performing models, we selected $N$ top-performing large-scale models as our teacher models. denoted by $\{A_i(\cdot)\}_{i=1}^N$. The detailed solution $y_{seed_{i}}^{*}$ generated by teacher model $A_{i}(\cdot)$ is calculated by:
\begin{equation}
    y_{seed_{i}}^{*}=\left\{
	\begin{array}{rcl}
		A_{i}(T_{detail}((x_{seed}, y_{seed}))) & & \text{ if not detailed}\\
		y_{seed} & & \text{if detailed}\\
	\end{array} \right.
\end{equation}
The prompt template $T_{detail}$ is shown on Appendix~\ref{Detail Prompt Template}.

\subsubsection{Knowledge \& Skill Extraction and Summary}
With $(x_{seed}, y_{seed}^{*})$, the domain knowledge points and problem-solving skills examined by $x_{seed}$, denoted as $K$, in each teacher model's perspective could be extracted using the prompt template $T_{extract}$ (see Appendix~\ref{Extract Prompt Template}):

Given that training techniques and data processing methods vary across different models, even high-performing teacher models may exhibit disparities in analytical capabilities. As each teacher model could generate biased or partial knowledge points and problem-solving techniques, it is essential to validate and summarize the insights derived from multiple teacher models. This integration process can be performed by any one of the teacher models.

\subsubsection{Training Data Generation}
Each teacher model $A_{i}(\cdot)$ is required to generate new examination questions $(x_{train_{i}}, y_{train_{i}})$ based on its own understanding of knowledge points and problem-solving methodologies integrated and summarized.
\begin{equation}
	(x_{train_{i}}, y_{train_{i}})=A_{i}(T_{generate}(K))
\end{equation}
where the generation prompt template $T_{generate}$ is shown on Appendix~\ref{Generate Prompt Template}.

\subsubsection{Peer-Review-Based Data Filtering}
\label{peer review}
Since the new training samples are entirely synthesized by large language models, which are prone to generating hallucinations and inaccuracies, it is necessary to implement a correctness filtering process for all artificially generated data. After the initial data generation phase using multiple teacher models, we implement a robust filtering mechanism inspired by the academic peer-review process \cite{manakul2023selfcheckgpt}. 

Given the generated data $(x_{train_{i}}, y_{train_{i}})$, the model $A_{i}(\cdot)$ that generated $y_{train_{i}}$ is designated as the \textit{candidate}. The remaining models $\mathcal{T}_{-i} = \mathcal{T} \setminus \{T_i\}$ act as the \textit{reviewer committee}. Each reviewer model $T_j \in \mathcal{T}_{-i}$ evaluates the candidate's answer $y_{train_{i}}$ against the generated question $x_{train_{i}}$. This evaluation produces a credibility score $s_{j \rightarrow i}$, which quantifies the extent to which the reviewer deems the candidate's answer to be correct and trustworthy. 

The credibility score $s_{j \rightarrow i}$ is generated by verification scoring: the reviewer is prompted to generate a probability score, which is normalized to a scalar value $s_{j \rightarrow i} \in [0, 1]$ (The prompt is shown in Appendix~\ref{peer review prompt}). The individual reviewer scores are aggregated into a final peer-review consensus score, $S_i$, for the candidate answer $y_{train_{i}}$. We employ a mean aggregation policy for its robustness:
\begin{equation}
	S_i = \frac{1}{|\mathcal{T}_{-i}|} \sum_{T_j \in \mathcal{T}_{-i}} s_{j \rightarrow i}
\end{equation}
A pre-defined confidence threshold \( \tau \in [0, 1] \) is then applied to make a binary decision on the inclusion of the data point in the final, curated dataset $\mathcal{D}_{reviewed}$. The decision rule is:
\begin{equation}
	(x_{train_{i}}, y_{train_{i}}) \rightarrow \mathcal{D}_{reviewed} \iff S_i \geq \tau
\end{equation}

This process is repeated for every triplet $(x_{train_{i}}, y_{train_{i}})$ in the raw dataset \( \mathcal{D}_{raw} \) with its generator $A_i(.)$. The final, high-quality dataset is therefore defined as:
\begin{equation}
	\begin{aligned}
		\mathcal{D}_{reviewed} &= \\
		&\{(x_{train_{i}}, y_{train_{i}}) \mid (x_{train_{i}}, y_{train_{i}})\\ &\in \mathcal{D}_{raw} \land S_i \geq \tau \}
	\end{aligned}	
\end{equation}

\subsection{Aligning: Data Trajectory Alignment}
\subsubsection{Summarize the Student Model’s Answer Style} Given training dataset $\mathcal{D}_{reviewed}$ and student LLM  $f(.)$, the student model answering set $\mathcal{D}_{\mathrm{style}}$ could be generated by
\begin{equation}
	\begin{aligned}
		\mathcal{D}_{\mathrm{style}}=\{(x, f(x)) \mid (x,y)\in \mathcal{D}_{\mathrm{reviewed}}\}
	\end{aligned}	
\end{equation}
The answer style of student LLM denoted as $S_{a}$, including language, tone, formatting structure and details, is summarized from $\mathcal{D}_{\mathrm{style}}$ by a SOTA model using prompt template in Appendix~\ref{Summarize Prompt Template}

\subsubsection{Construct Trajectory Aligned Data}
The trajectory aligned solution of original question $x_{train}$ with respect to student model $f(.)$, denoted as $y_{aligned_i}$, could be generated by teacher model $A_i(.)$ with the alignment prompt $T_{align}$, by the equation 
\begin{equation}
	y_{aligned_i}=A_{i}(T_{align}(S_{a}, x_{train}, y_{train}))
\end{equation}
where $T_{align}$ is formatted by the template in Appendix~\ref{Align Prompt Template}.

\subsubsection{Candidates Reflection \& Voting}
For each $x_{train}$, we obtain a candidate set $\mathcal{Y}(x_{train})=\{y_{aligned_i}\}_{i=1}^{N}$.
We rank candidates by a sum of standardized student-informativeness and reward scores:
\[
\mathcal{R}_{\mathrm{total},i}=\text{z-score}_{x}\!\left(\mathcal{R}_{\mathrm{student},i}\right)+
\text{z-score}_{x}\!\left(\mathcal{R}_{\mathrm{reward},i}\right),
\]
where standardization is performed \emph{within the candidate set of the same $x$}.
The best candidate is $y_{aligned_i}^{*}=\arg\max_{i}\mathcal{R}_{\mathrm{total},i}$. The aligned training dataset is
\begin{equation}
	\mathcal{D}_{aligned} = \{(x_{train}, y_{aligned_i}^{*})\}
\end{equation}

\paragraph{Reflection on Student model} The reversed-IFD score published by Selective Reflection-Tuning \cite{li-etal-2024-selective} provides a criterion to hypothesize that a response is more informative for training if it is feasible for the LLM to predict the corresponding instruction from the response. Encouraged by the hypothesis, we  use the r-IFD equation in \cite{li-etal-2024-selective} as the student-reflection term
$$ \mathcal{R}_{student_i} =-\operatorname{log}(\frac{\operatorname{ppl}(x_{train}|y_{aligned_i}^{'})}{\operatorname{ppl}(x_{train})}) $$
where $y_{aligned_i}^{'}$ represents the text segment generated by mapping the original $y_{aligned_i}$ into a query to guess the
corresponding potential instructions.
\paragraph{Reflection on Reward Model} A judge model assesses each candidate answer based on a predefined set of criteria.
The judge model assigns a score $s_{i,k}$ to candidate answer $y_{aligned_i}$ for criterion $c_k$, producing a score vector:
\begin{equation}
	\mathbf{s}_i = [s_{i,1}, s_{i,2}, \dots, s_{i,K}]^T
\end{equation}
To aggregate these scores into a single ranking, let $\mathbf{w} = [w_1, w_2, \dots, w_K] $ be a weight vector representing the relative importance of each criterion, where $\sum_{k=1}^K w_k = 1$. The overall quality score for answer $y_{aligned_i}$ is computed as:
\begin{equation}
 \mathcal{R}_{reward_i} = \mathbf{w}^T \mathbf{s}_i = \sum_{k=1}^K w_k \cdot s_{i,k}
\end{equation}
In practical, the criteria include:
\begin{itemize}
	\item \textbf{Correctness:} Factual accuracy and freedom from hallucinations.
	\item \textbf{Completeness:} Thoroughness in addressing all aspects of query.
	\item \textbf{Clarity:} Coherence, fluency, and organizational structure.
	\item \textbf{Conciseness:} Brevity and absence of redundant information.
\end{itemize}
The reward prompt is shown in Appendix~\ref{reflection prompt}. The overall reward weight $(w_{\mathrm{corr}},w_{\mathrm{comp}},w_{\mathrm{clar}},w_{\mathrm{conc}})=(0.5,0.2,0.2,0.1)$.


\section{Experiments}
\subsection{Experiment Setup}
\subsubsection{Scope}
We restrict our study to \emph{supervised fine-tuning (SFT)} with cross-entropy.
All systems share the same student architecture, optimizer, and decoding setup.
Comparisons are made under an \emph{equal training-token budget}.
We do not train RLHF/DPO/PRM/verifier models, nor do we employ tool-use or retrieval during training.

Baselines are SFT-only and differ only by data construction or scheduling. Unless noted, test-time reasoning is disabled.
\subsubsection{Model Selection}
We adopt \textsc{Qwen3-32B} as the student (base) model. The teacher ensemble comprises \textsc{DeepSeekV3}, \textsc{Qwen3-235B}, and \textsc{Kimi K2}. We use \textsc{GPT-5} both as the style summarization model (for DTA style induction) and as the reflection-based reward judge \cite{deepseekv3-techreport, qwen3-techreport, kimi-k2-techreport}. Unless otherwise noted, all inference in \textbf{Phase I: Initializing} and \textbf{Phase II: DTA} are executed on NVIDIA A800 GPUs with \texttt{vLLM==0.9.1} \cite{kwon2023vllm}.

\subsubsection{Training Data Statistics}
We use the original \textsc{Telemath} benchmark \cite{colle2025telemathbenchmarklargelanguage} as seed data. Each teacher model generates 5{,}000 candidate training examples, yielding a raw pool that is filtered via peer review in Section~\ref{peer review}. 

Then, to mitigate potential data leakage arising from our use of the benchmark as seed material for distillation, we implemented a MinHash-based \cite{broder1997minhash} decontamination pipeline. Specifically, we computed k-shingle fingerprints for every candidate training instance and for all benchmark items, applied locality-sensitive hashing (LSH \cite{indyk1998lsh}) to retrieve near-duplicate pairs, and estimated Jaccard similarity from the MinHash signatures. Training samples whose estimated similarity to any benchmark instance exceeded a conservative threshold were flagged and removed prior to model training. This procedure reduces contamination risk, prevents evaluation bias, and ensures that performance gains cannot be attributed to memorization of benchmark content, finally forming the curated set $\mathcal{D}_{reviewed}$. In practice, we compute MinHash signatures over case-folded, punctuation-stripped $k$-shingles (word-level, $k{=}5$ unless otherwise noted), using $H{=}128$ hash functions and an LSH scheme with $(b{=}32,\ r{=}4)$. Candidate train items are compared against all benchmark items; pairs with estimated Jaccard similarity $\ge \tau_{\mathrm{jac}}{=}0.8$ are removed. 

Beyond lexical near-duplicate removal with MinHash, we apply a semantic-level filter to catch paraphrases and algebraically equivalent instances that survive shingling. The pipeline has three stages. 
(i) \emph{Canonicalization.} We normalize casing, whitespace, and numeric formats; convert units to SI (e.g., dBm$\leftrightarrow$mW) via deterministic rules; sort key--value parameter lists by key; and canonicalize equations using symbolic algebra \cite{meurer2017sympy} (commutativity/associativity reordering, factorization, constant folding) to produce a stable prefix-string for formulas. 
(ii) \emph{Dual-view embedding retrieval.} For each candidate $(x_{\text{train}},y_{\text{train}})$ and each benchmark item, we compute dense embeddings for two complementary views: a text view $e_{\text{txt}}$ over the concatenated question and prose solution, and a structure view $e_{\text{form}}$ over the canonicalized formula/units string. We build an approximate nearest-neighbor index and retrieve top-$K$ neighbors by cosine similarity in each view (default $K{=}50$). A pair enters the verification stage if $\max\{\cos(e_{\text{txt}}^{(c)},e_{\text{txt}}^{(b)}),\ \cos(e_{\text{form}}^{(c)},e_{\text{form}}^{(b)})\}\ge \tau$ with view-specific thresholds ($\tau_{\text{txt}}{=}0.86$, $\tau_{\text{form}}{=}0.90$). 
(iii) \emph{Cross-verification.} For each retrieved pair we run a lightweight cross-encoder/NLI-style judge to estimate bidirectional entailment $s_{\text{xenc}}$ between the two Q\&A items, and (when executable) a symbolic check that samples domain-valid numeric instantiations and evaluates both solutions, declaring equivalence if relative error $\le \epsilon$ (default $\epsilon{=}10^{-3}$). A candidate is flagged as contaminated iff the similarity sieve passes and either $s_{\text{xenc}}\ge\tau_{\text{xenc}}$ (default $0.60$) \emph{or} the symbolic/numeric check confirms equivalence. We also perform template slot-abstraction (variable renaming, unit-conversion patterns) to catch re-parameterized clones. 
This semantic filter, applied \emph{after} MinHash and \emph{before} training, removes paraphrases, parameter renamings, and unit-conversion duplicates that would otherwise inflate accuracy; thresholds were selected via grid search on a small held-out slice of \textsc{Telemath} dev pairs. The final curated set size used for training is as reported above.

Finally, the training set contains 14{,}190 samples. The sample removal summary is show in \ref{data_removal_per_teacher}. This set is used in two ways: (i) directly for the distilled-only baseline and (ii) as input to DTA to produce the aligned corpus $\mathcal{D}_{aligned}$.

\begin{table}[t]
	\centering
	\caption{Sample Removal Summary for Each Teacher Model in Data Curation Process}
	\label{data_removal_per_teacher}
	\begin{tabular}{lccc}
		\toprule
		\textbf{Method} & \textbf{DeepSeek-V3} & \textbf{Qwen3 235B} & \textbf{Kimi-K2} \\
		\midrule
		Peer Review & 9 & 0 & 14 \\
		MinHash & 58 & 72 & 82 \\
		Semantic & 169 & 189 & 217 \\
		\bottomrule
	\end{tabular}
\end{table}

\subsubsection{Training Setting}
\label{training_setting}
We fine-tune with identical hyperparameters for both $\mathcal{D}_{reviewed}$ (Vanilla-Distill) and $\mathcal{D}_{aligned}$ (Distill + DTA) to enable controlled comparisons.

\begin{table}[h]
	\centering
	\caption{Training hyperparameters (held fixed across all runs).}
	\label{tab:hyper}
	\begin{tabular}{cc}
		\toprule
		Hyperparameter & Value \\
		\midrule
		Batch size & 64 \\
		Learning rate & $1\times 10^{-6}$ \\
		LR schedule & constant \\
		Weight decay & $1\times 10^{-1}$ \\
		Epochs & 3 \\
		Warmup ratio & 0 \\
		AdamW $\beta_{1}$ & 0.95 \\
		AdamW $\beta_{2}$ & 0.95 \\
		Gradient clipping & 1.0 \\
		Max seq len & 8192 \\
		\bottomrule
	\end{tabular}
\end{table}
We train \textsc{Qwen3-32B} with \emph{thinking mode disabled} on $\mathcal{D}_{reviewed}$ and $\mathcal{D}_{aligned}$ separately using the configuration in Table~\ref{tab:hyper} with the hardware of Ascend 910B3 with Mindspeed-LLM framework. The resulting models are denoted \textbf{$g1_{\text{tele}}$} (trained on $\mathcal{D}_{reviewed}$) and \textbf{$g2_{\text{tele}}$} (trained on $\mathcal{D}_{aligned}$).

\subsubsection{Inference Setting}
\label{inference_setting}
We evaluate inference efficiency and accuracy under edge-friendly settings:
\begin{itemize}
	\item \textbf{Hardware/runtime.} (a) NVIDIA A800 GPUs with \texttt{vLLM==0.9.1}; (b) KV-cache enabled; (c) batch size $=$ 8; (d) no dynamic batching; (e) retrieval/tools disabled at inference.
	\item \textbf{Tasks.} \textsc{Telemath}: (i) AMC selection under SNR constraints; (ii) power-control with max-EIRP; (iii) link-budget feasibility checks.
\end{itemize}
All efficiency metrics are averaged over 5 runs and reported as mean $\pm$ 95\% CI to smooth occasional GPU-side numeric fluctuations.

We use deterministic decoding for pass@1 and majority voting over 16 generated answers (cons@16) \cite{wang2022selfconsistency}:
\begin{table}[h]
	\centering
	\caption{Inference parameters for pass@1}
	\label{tab:pass1}
	\begin{tabular}{cc}
		\toprule
		parameter & value \\
		\midrule
		temperature & 0 \\
		top\_p & 0.01 \\
		max\_tokens & 32768 \\
		\bottomrule
	\end{tabular}
\end{table}

\begin{table}[h]
	\centering
	\caption{Inference parameters for cons@16}
	\label{tab:cons16}
	\begin{tabular}{cc}
		\toprule
		parameter & value \\
		\midrule
		temperature & 0.6 \\
		top\_k & 20 \\
		top\_p & 0.95 \\
		max\_tokens & 32768 \\
		\bottomrule
	\end{tabular}
\end{table}

We explicitly control thinking mode for evaluation via the vLLM chat template configuration, \emph{independent of prompting}. The request body includes:
\begin{verbatim}
	"chat_template_kwargs": {
		"enable_thinking": true  // or false
	}
\end{verbatim}

\subsection{Experiment Result}
\subsubsection{Overall}
\begin{table*}[t]
	\centering
	\caption{The leaderboard of \textsc{Telemath}, \textbf{pass@1} is the ability to generate a correct answer in a single attempt and \textbf{cons@16} means correctness based on majority voting over 16 generated answers}
	\label{tab:leaderboard}
	\begin{tabular}{cccc}
		\toprule 
		model& thinking mode enabled & Overall pass@1 & Overall cons@16 \\
		\midrule
		$g2_{\text{tele}}$ & No & \textbf{72.45}\% & \textbf{78.02}\% \\
		Qwen3-32B & Yes & 69.51\% & 76.00\% \\
		$g1_{\text{tele}}$ & No & 54.80\% & 61.64\% \\
		DeepSeek-R1-Distill-Llama-70B & Yes & 53.21\% & 60.80\% \\
		Phi-4-Reasoning+ & Yes & 53.56\% & 58.40\% \\
		Qwen3-32B & No & 47.80\% & 52.64\% \\
		Qwen3-4B & Yes &45.62\% & 50.00\% \\
		Qwen2.5-Math-72B-Instruct & No & 39.99\% & 46.48\% \\
		Llama-3.3-70B-Instruct & No & 36.23\% & 40.20\% \\
		Qwen2.5-Math-7B-Instruct & No & 22.38\% & 26.60\% \\
		Llama-3.1-8B-Instruct & No & 13.56\% & 20.00\% \\
		\bottomrule
	\end{tabular}
\end{table*}
	Table~\ref{tab:leaderboard}  presents the performance of 8 representative general LLMs and our telecommunication math expert LLM $g1_{\text{tele}}$ and $g2_{\text{tele}}$, on the \textsc{Telemath} benchmark. The models are ranked in descending order by their overall scores. 
	It shows that the DTA-refined model $g2_{\text{tele}}$ attains the best results on Overall pass@1 and Overall cons@16 despite not enabling an explicit thinking mode. Concretely, $g2_{\text{tele}}$ reaches \textbf{72.45\%} pass@1 and \textbf{78.02\%} cons@16, surpassing both its distilled-only counterpart $g1_{\text{tele}}$ and the strong baseline Qwen3-32B with thinking mode enabled.
	
	Relative to $g1_{\text{tele}}$ (54.80\% pass@1, 61.64\% cons@16), $g2_{\text{tele}}$ improves by
	$\Delta\text{pass@1}=+17.65$ points (relative +32.2\%)
	and $\Delta\text{cons@16}=+16.38$ points (relative +26.6\%).
	These gains indicate that DTA in data curation is a dominant factor beyond simple distillation: it lifts one-shot accuracy and robustness under self-consistency, even when the base model does not use an explicit thinking mode.
	
	Compared to Qwen3-32B with thinking mode enabled (69.51\% pass@1, 76.00\% cons@16), $g2_{\text{tele}}$ yields
	$+2.94$ and $+2.02$ absolute points on pass@1 and cons@16, respectively (relative +4.2\% and +2.7\%). 
	Thus, DTA-trained supervision closes and slightly overturns the gap to a larger, thinking-enabled baseline—\emph{without} relying on inference-time reasoning heuristics.
	
	Against Qwen3-32B without thinking mode (47.80\% / 52.64\%), $g2_{\text{tele}}$ delivers large improvements of
	$+24.65$ (pass@1) and $+25.38$ (cons@16), i.e., relative +51.6\% and +48.2\%.
	This highlights the importance of high-signal, trajectory-aligned supervision over raw scale or generic instruction tuning when adapting to telecommunications mathematics.
	
	The pass@1$\to$cons@16 lift for $g2_{\text{tele}}$ is 5.57 points (78.02$-$72.45), which is smaller than for $g1_{\text{tele}}$ (6.84) and  Qwen3-32B with thinking mode (6.49).
	This suggests that DTA reduces reliance on multi-sample voting by yielding more deterministically correct first-pass solutions—useful when latency or sampling budgets are constrained.
	
	Across both accuracy (pass@1) and robustness (cons@16), DTA transforms distilled data into compact, high-yield supervision that (i) substantially outperforms distilled-only training and (ii) edges out a strong thinking-enabled baseline, all while operating without an explicit thinking mode. These results support the central claim that \emph{aligning solution trajectories and answer style with the target student model} is a highly effective—and inference-efficient—lever for domain adaptation in low-resource verticals.


\subsubsection{Token Shift Analysis}
\begin{figure}[t]
	\centering
	\subfigure[$g1_{tele}$]{\includegraphics[width=0.2\textwidth]{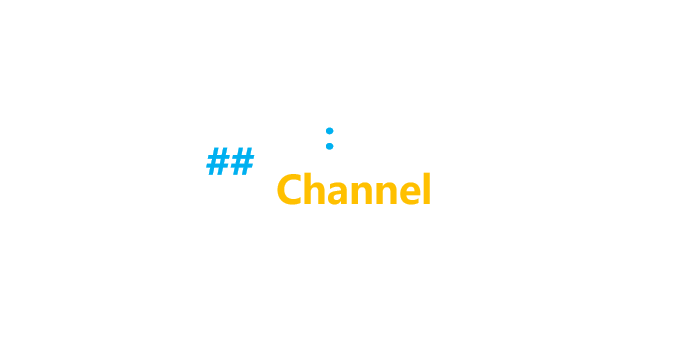}}
	\subfigure[$g2_{tele}$]{\includegraphics[width=0.2\textwidth]{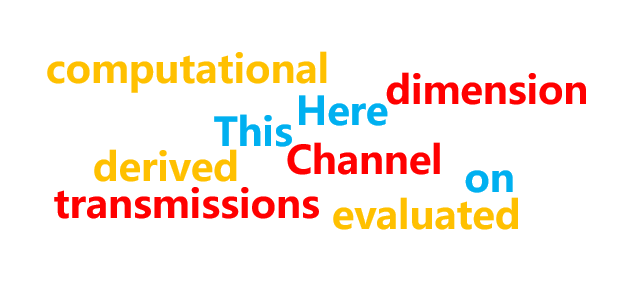}}
	\caption{Word cloud of token shift in \textsc{Telemath} benchmark}
	\label{token shift telemath}	
\end{figure}

Figure~\ref{token shift telemath} visualizes token-level shifts \cite{lin2023unlockingspellbasellms} between the distilled-only model $g1_{tele}$ and the DTA-refined model $g2_{\text{tele}}$ on the \textsc{Telemath} benchmarks. Tokens are extracted by comparing frequency and rank against Qwen3-32B without explicit thinking mode and are categorized as: (i) \emph{logical–structural} discourse markers (highlighted in {red}), (ii) \emph{telecommunications domain-specific} terms (in {orange}), and (iii) \emph{content-specific} nouns and numerals (in {blue}). Two consistent patterns emerge from the word clouds:
\begin{enumerate}
	\item \textbf{DTA concentrates shifts on logical scaffolding.} In $g2_{tele}$, the majority of high-magnitude shifts comes from logical–structural tokens (e.g., derived, evaluated). This indicates a stronger adoption of explicit discourse structure and step demarcation in generated solutions.
	\item \textbf{Distilled-only training over-weights content tokens.} In $g1_{tele}$, none of the top-shifted tokens are logical–structural; the cloud skews toward content nouns, numerals, and symbol-like tokens. This suggests a tendency to recall topical or formulaic surface forms rather than to externalize the reasoning glue that connects steps.
\end{enumerate}

The relative dominance of logical–structural markers in $g2_{tele}$ is a direct signature of trajectory alignment: DTA does not merely change which facts are produced, but regularizes how solutions are narrated and decomposed. In the context of telecommunications mathematics—where problems interleave algebraic manipulation with engineering constraints—explicit discourse cues (contrast, implication, sequencing) serve as a lightweight control flow that (i) stabilizes intermediate calculations, (ii) reduces local ambiguities when propagating constraints, and (iii) promotes verifiable step-by-step justifications. By contrast, the content-centric shifts of $g1_{tele}$ imply heavier reliance on topical tokens and numeric fragments, which offers less guardrail against error propagation when intermediate steps are underspecified.

The token-shift profile of $g2_{tele}$ aligns with its empirical advantages on \textsc{Telemath} (higher pass@1 and cons@16 compared to both $g1_{tele}$ and Qwen3-32B without thinking). Concentrating shift mass on discourse markers is consistent with producing more deterministically correct first-pass solutions—i.e., fewer cases where correctness depends on multi-sample voting—while retaining the necessary domain terminology to satisfy telecommunication-specific constraints. In short, DTA steers the model toward reasoning \emph{process} tokens, not just domain content tokens, and this shift corresponds to measurable gains in accuracy and consistency.

\subsection{Mobile/Edge Measurement Analysis}
\label{sec:mobile}
\textbf{Goal:}
We quantify the system-level impact of DTA under edge-like constraints where compute, memory bandwidth, and energy are tight. We focus on \emph{first-pass} efficiency (i.e., no re-sampling or majority voting) because improving pass@1 directly reduces latency/energy budgets in practical deployments.

\textbf{Testbed and protocol:} Inference settings are in Section~\ref{inference_setting}. We evaluate four models: \textbf{$g2_{\text{tele}}$} (DTA), \textbf{$g1_{\text{tele}}$} (distilled-only), and the base \textbf{Qwen3-32B} with/without thinking.

\textbf{Primary metrics:}
\begin{itemize}
	\item \textbf{Energy per token} (J/token): sampled via \texttt{nvidia-smi dmon} at 10\,Hz and normalized by generated \emph{output} tokens.\vspace{-2pt}
	\item \textbf{Latency} (s/sample): enqueue $\rightarrow$ last-token time.\vspace{-2pt}
	\item \textbf{Throughput} (tokens/s): steady-state decode rate reported by the server.
\end{itemize}

\begin{table}[t]
	\centering
	\caption{Latency, throughput, and energy measurements}
	\label{mobile}
	\begin{tabular}{ccccc}
		\toprule 
		model & thinking & Energy $\downarrow$ & Latency $\downarrow$ & Throughput $\uparrow$ \\
		\midrule
		$g2_{\text{tele}}$ & No  & 0.96 & 2.87 & 310.99 \\
		Qwen3-32B          & Yes & 1.65 & 45.89 & 181.39 \\
		$g1_{\text{tele}}$ & No  & 1.19 & 2.52 & 251.74 \\
		Qwen3-32B          & No  & 1.60 & 7.18 & 187.39 \\
		\bottomrule
	\end{tabular}
	\vspace{-6pt}
\end{table}

To connect systems metrics with task success, we also report:
\begin{itemize}
	\item $\text{Tokens/sample} = \text{Throughput}\times\text{Latency}$
	\item $\text{Energy/sample} = (\text{J/token}) \times \text{Tokens/sample}$
	\item $\text{Energy–Delay Product (EDP)}^* = (\text{J/token}) \times \text{Latency}$, lower is better
	\item $\text{Time per correct answer} = \frac{\text{Latency}}{\text{pass@1}}$
	\item $\text{Energy per correct answer} = \frac{\text{Energy/sample}}{\text{pass@1}}$, where pass@1 comes from Table~\ref{tab:leaderboard}.
\end{itemize}
Because outputs are of similar structure across non-thinking systems, these proxies are informative for relative comparisons (we note limitations below).

\begin{table*}[t]
	\centering
	\caption{Derived efficiency metrics (lower is better unless noted).}
	\label{tab:mobile_derived}
	\begin{tabular}{lccccc}
		\toprule
		model & Tokens/sample $\downarrow$ & EDP$^*$ (J$\cdot$s/token) $\downarrow$ & Energy/sample (J) $\downarrow$ & Time/correct (s) $\downarrow$ & Energy/correct (J) $\downarrow$ \\
		\midrule
		$g2_{\text{tele}}$ & 892.5 & 2.755 & 856.8 & \textbf{3.96} & \textbf{1182.7} \\
		Qwen3-32B (Yes)    & 8324.0 & 75.718 & 13734.6 & 66.02 & 19759.1 \\
		$g1_{\text{tele}}$ & \textbf{634.4} & 2.999 & \textbf{754.9} & 4.60 & 1377.6 \\
		Qwen3-32B (No)     & 1345.5 & 11.488 & 2152.7 & 15.02 & 4503.6 \\
		\bottomrule
	\end{tabular}
	\vspace{-6pt}
\end{table*}

The key findings are:
\begin{itemize}
	\item \textbf{DTA vs. thinking baseline.} Relative to Qwen3-32B (thinking enabled), $g2_{\text{tele}}$ reduces energy per token by \textbf{41.8\%}, improves throughput by \textbf{71.4\%}, and cuts latency by \textbf{93.7\%}. EDP$^*$ drops by \textbf{96.4\%}. Accounting for accuracy, \emph{time per correct} shrinks by \textbf{94.0\%} and \emph{energy per correct} by \textbf{94.0\%}. The dominant factor is the massive \emph{tokens/sample} explosion under thinking mode (8.3k vs. 0.89k).
	\item \textbf{DTA vs. base (thinking mode disabled).} Against Qwen3-32B (thinking mode disabled), $g2_{\text{tele}}$ lowers energy per token by \textbf{40.0\%} and latency by \textbf{60.0\%}, with \textbf{+66.0\%} throughput. EDP$^*$ falls by \textbf{76.0\%}; \emph{time/energy per correct} improve by \textbf{73.6\%}/\textbf{73.7\%}. 
	\item \textbf{DTA vs. distilled-only.} Versus $g1_{\text{tele}}$, DTA reduces energy per token by \textbf{19.3\%} and increases throughput by \textbf{23.5\%}. Although $g2_{\text{tele}}$ emits longer answers (892 vs. 634 tokens/sample), its higher pass@1 (\S\ref{tab:leaderboard}) makes \emph{time per correct} \textbf{13.9\%} lower and \emph{energy per correct} \textbf{14.1\%} lower than $g1_{\text{tele}}$.
\end{itemize}

In brief, DTA aligns intermediate steps and formatting with the student, which (i) raises first-pass correctness, reducing the need for re-sampling/majority voting; (ii) avoids expensive “thinking” tokens while preserving the \emph{useful} scaffolding (token-shift analysis in Figure~\ref{token shift telemath}); and (iii) stabilizes unit/constraint checks so decoding does not meander (higher tokens/s, fewer backtracks). 
In short, DTA moves cost from inference-time \emph{sampling} to training-time \emph{curation}, which is favorable for mobile/edge deployments with the following practical guidance:
\begin{itemize}
	\item \textbf{Prefer DTA-curated, non-thinking decoding} for edge: it delivers the best \emph{time/energy per correct} without chain-of-thought overhead.
	\item \textbf{Tune output budgets, not temperature.} Since $g2_{\text{tele}}$ already improves pass@1, capping max new tokens (or adding concise-style constraints) yields further latency/energy wins with minor accuracy impact.
	\item \textbf{Exploit KV-cache and steady-state decode.} Gains are decode-heavy; keeping prompts compact and reusing system prompts magnifies the tokens/s advantage of DTA-trained models.
\end{itemize}

In summary, DTA improves \emph{first-pass} accuracy and decode efficiency simultaneously, enabling telecom-grade reasoning \emph{without} thinking-mode overhead. The resulting $\sim$2--4$\times$ improvements in \emph{time/energy per correct answer} versus strong non-thinking baselines---and $\sim$16--20$\times$ versus thinking-mode---make DTA a practical fit for mobile/edge deployments where latency and energy are first-class constraints.

\section{Ablation: Generation-Free DTA on General Mathematics}
To isolate the effect of DTA from teacher-driven data distillation, we conduct a distillation-free ablation on general mathematics.
Instead of generating supervision with a high-capacity teacher, we operate solely on publicly available math corpora $\mathcal{D}_{math}$ and apply direct DTA.
The aligned set $\mathcal{D}_{math\_aligned}$ preserves gold final answers and edits only intermediate rationales; no external distillation is queried during construction. We compare 2 training conditions under \emph{identical} architectures, optimizers, learning-rate schedules, context lengths, and total training-token budget $B$ (matched by subsampling): (i) \textbf{Raw} (no DTA) using $\mathcal{D}_{math}$; and (ii) \textbf{DTA-only} using $\mathcal{D}_{math\_aligned}$.

\subsection{Experiment Setup}
\subsubsection{Model Selection}
We adopt \textsc{Qwen3-32B} as the student (base) model. The teacher ensemble comprises \textsc{DeepSeekV3}, \textsc{Qwen3-235B}, and \textsc{Kimi K2}. These models are used for style induction and trajectory rewriting within DTA; no changes are made to the student architecture across conditions.

\subsubsection{$\mathcal{D}_{\text{math}}$}
We construct the math corpus from two top-ranking, open-source training datasets for the Qwen3 series on the \textsc{OpenDataArena} platform~\cite{opendataarena_tool_2025}: \textsc{Light-R1-SFTData}~\cite{wen2025lightr1curriculumsftdpo} and \textsc{OpenThoughts-114k}~\cite{guha2025openthoughtsdatarecipesreasoning}. These datasets are used (i) as a raw baseline and (ii) as inputs to produce a trajectory-aligned variant $\mathcal{D}_{\text{math\_aligned}}$ via DTA.

\subsubsection{Training Setting}
Unless otherwise specified, the training configuration (e.g. optimizer, learning-rate schedule, batch size, weight decay, epochs) exactly matches Section~\ref{training_setting}. 

We fine-tune \textsc{Qwen3-32B} with \emph{thinking mode disabled} on $\mathcal{D}_{\text{math}}$ and $\mathcal{D}_{\text{math\_aligned}}$ separately, using the hyperparameters in Section~\ref{training_setting}. The resulting models are denoted \textbf{$g1_{\text{gene}}$} (trained on $\mathcal{D}_{\text{math}}$) and \textbf{$g2_{\text{gene}}$} (trained on $\mathcal{D}_{\text{math\_aligned}}$).

\subsubsection{Inference Setting}
We evaluate inference accuracy under edge-friendly settings:
\begin{itemize}
	\item \textbf{Hardware/runtime.} the same as Section~\ref{inference_setting}
	\item \textbf{Tasks.} \textsc{AoPS2024}~\cite{mahdavi2025leveragingonlineolympiadlevelmath} and \textsc{OlympiadBench}~\cite{he2024olympiadbenchchallengingbenchmarkpromoting}
\end{itemize}
All efficiency metrics are averaged over 5 runs and reported as mean $\pm$ 95\% CI to smooth occasional GPU-side numeric fluctuations.

We use deterministic decoding for pass@1 under the same parameters with \ref{inference_setting}.

\subsection{Experiment Result}
\subsubsection{Overall}
\begin{table}[t]
	\centering
	\caption{Leaderboard on general mathematics. All results are pass@1 with \emph{thinking disabled} at inference.}
	\label{math_result}
	\begin{tabular}{cccc}
		\toprule 
		\textbf{model} & \textbf{thinking mode enabled} & \textbf{AoPS2024} & \textbf{OlympiadBench} \\
		\midrule
		$g2_{\text{gene}}$ & No & \textbf{38.3\%} & \textbf{50.8\%} \\
		$g1_{\text{gene}}$ & No & 33.3\% & 48.3\% \\
		\textsc{Qwen3-32B} & No & 36.3\% & 49.5\% \\
		\bottomrule
	\end{tabular}
\end{table}

\begin{figure}[t]
	\centering
	\subfigure[$g2_{\text{gene}}$ vs.\ $g1_{\text{gene}}$]{\includegraphics[width=0.4\textwidth]{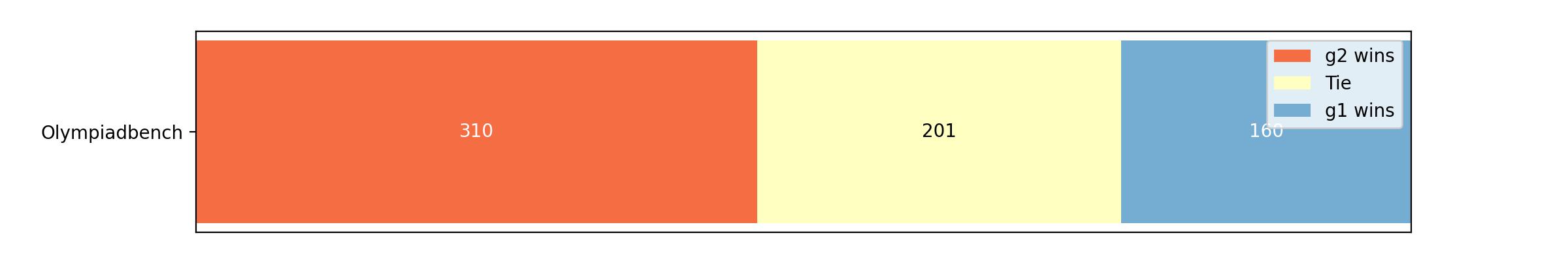}}
	\subfigure[\textsc{Qwen3-32B} vs.\ $g1_{\text{gene}}$]{\includegraphics[width=0.4\textwidth]{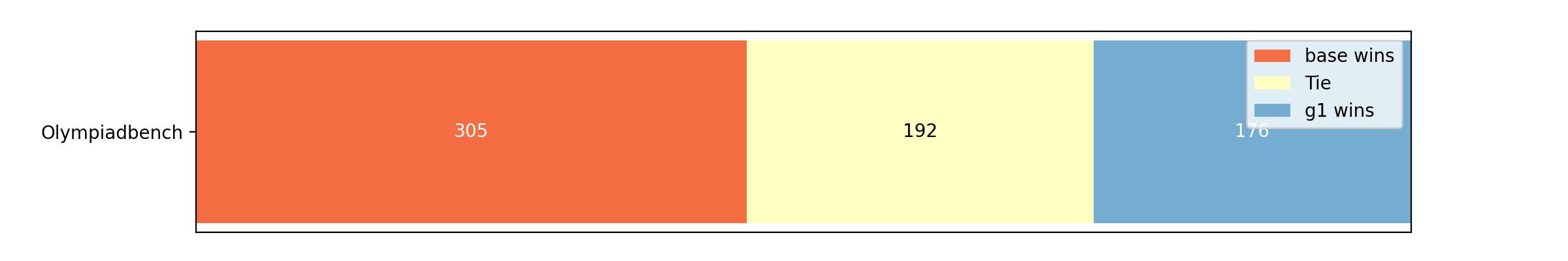}}
	\subfigure[\textsc{Qwen3-32B} vs.\ $g2_{\text{gene}}$]{\includegraphics[width=0.4\textwidth]{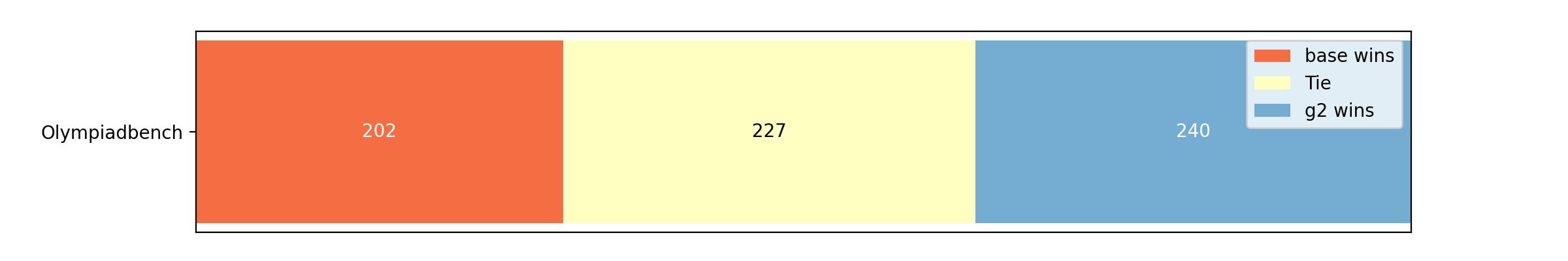}}
	\caption{Pairwise LLM-judge comparisons between models on \textsc{OlympiadBench}.}
	\label{comparison}	
\end{figure}

Table~\ref{math_result} compares the base model (\textsc{Qwen3-32B}) with two fine-tuned variants on \textsc{AoPS2024}~\cite{mahdavi2025leveragingonlineolympiadlevelmath} and \textsc{OlympiadBench}~\cite{he2024olympiadbenchchallengingbenchmarkpromoting}. 

To ensure the robustness and statistical significance of the results, we conducted independent experiments with at least three different random seeds for each model. The Pass@1 accuracy of each model was computed, and the mean and 95\% confidence intervals (CI) were calculated.

For $g2_{\text{gene}}$, the mean Pass@1 accuracy on \textsc{AoPS2024} was \textbf{38.3\%}, with a 95\% CI of [37.4\%, 39.2\%], and on \textsc{OlympiadBench}, it was \textbf{50.8\%}, with a 95\% CI of [49.4\%, 52.2\%]. In comparison, Qwen3-32B (without thinking mode enabled) achieved a mean Pass@1 accuracy of \textbf{36.3\%} on \textsc{AoPS2024}, with a 95\% CI of [35.5\%, 37.1\%], and \textbf{49.5\%} on \textsc{OlympiadBench}, with a 95\% CI of [48.6\%, 50.4\%]. $g1_{\text{tele}}$, trained only on distilled data, achieved a lower mean Pass@1 accuracy of \textbf{33.3\%} on \textsc{AoPS2024}, with a 95\% CI of [32.5\%, 34.1\%], and \textbf{48.3\%} on \textsc{OlympiadBench}, with a 95\% CI of [47.2\%, 49.4\%]. The tight confidence intervals reflect the reliability of these performance gains.

The DTA-trained $g2_{\text{gene}}$ attains the best accuracy on both benchmarks, surpassing $g1_{\text{gene}}$ (raw SFT without trajectory alignment) by \textbf{+5.0} points on \textsc{AoPS2024} (33.3\% $\rightarrow$ 38.3\%; $\sim$15.0\% relative) and by \textbf{+2.5} points on \textsc{OlympiadBench} (48.3\% $\rightarrow$ 50.8\%; $\sim$5.2\% relative). Against the unfine-tuned base, $g2_{\text{gene}}$ yields absolute gains of \textbf{+2.0} (\textsc{AoPS2024}) and \textbf{+1.3} (\textsc{OlympiadBench}) (5.5\% and 2.6\% relative, respectively). Interpreted as error-rate reduction, DTA lowers mistakes by $\sim$7.5\% on \textsc{AoPS2024} and $\sim$4.8\% on \textsc{OlympiadBench} relative to $g1_{\text{gene}}$.

These improvements persist across two contest-style datasets with differing item mixes, indicating that DTA provides \emph{domain-general} benefits rather than overfitting to a specific distribution. Moreover, the strong zero-tuning performance of the base model underscores that the observed gains are not attributable to additional tokens or model scale; instead, they arise from \emph{how} supervision is curated—namely, aligning intermediate steps and presentation style prior to SFT.

Figure~\ref{comparison} reports LLM-judge head-to-head preferences (prompt template in Appendix~\ref{pairwise}) on \textsc{OlympiadBench}. Across matchups, $g2_{\text{gene}}$ is consistently preferred over $g1_{\text{gene}}$, and remains competitive with the base, corroborating the exact-match trends in Table~\ref{math_result}. The agreement between preference wins and accuracy suggests that DTA’s advantage is not an artifact of the judging protocol but reflects broader improvements in helpfulness, relevance, and correctness.

\subsubsection{Training Loss Comparison}
\label{sec:math_loss}
\begin{figure}[t]
	\centering
	\includegraphics[width=0.48\textwidth]{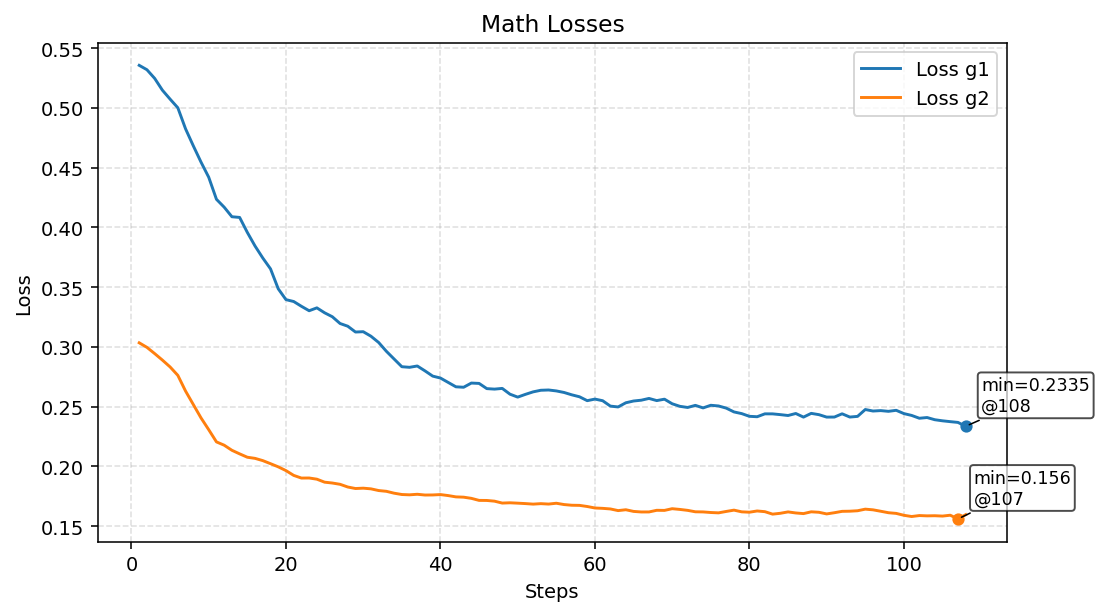}
	\caption{Training loss in the math ablation: $g2_{\text{gene}}$ (DTA-only) converges faster and to a lower minimum than $g1_{\text{gene}}$ (raw SFT).}
	\label{fig:math_loss}
\end{figure}

Figure~\ref{fig:math_loss} contrasts token-level training loss for \textbf{$g1_{\text{gene}}$} (raw SFT) and \textbf{$g2_{\text{gene}}$} (DTA-only) under identical optimization settings. Three observations follow:

\begin{itemize}
	\item \textbf{Lower initialization and faster early descent.} At the start of training, $g2_{\text{gene}}$ exhibits a markedly lower loss than $g1_{\text{gene}}$ (approximately 0.305 vs.\ 0.536; a \textbf{43\%} relative gap) and descends more steeply over the first 20--30 steps. This indicates that trajectory-aligned targets are easier to fit from the outset, reducing the gradient ``warm-up'' required to locate a favorable basin.
	\item \textbf{Better asymptote.} The minimum loss of $g2_{\text{gene}}$ is \textbf{0.156} (at step~107), versus \textbf{0.2335} for $g1_{\text{gene}}$ (step~108), a \textbf{33.2\%} reduction (\(\Delta L{=}0.0775\)). This mirrors the accuracy gains in Table~\ref{math_result} and suggests that the lower cross-entropy reflects cleaner, higher-signal supervision rather than overfitting.
	\item \textbf{Smoother convergence.} Beyond step~40, $g2_{\text{gene}}$ shows smaller oscillations and plateaus earlier, whereas $g1_{\text{gene}}$ exhibits several bumps (around steps 50--70 and 90--100). Reduced variance in the loss trajectory is consistent with less label/style noise and more stable gradients.
\end{itemize}

DTA narrows the mismatch between the model’s inductive biases and the supervision distribution. Concretely, rewriting intermediate steps into a consistent student style (i) \emph{lowers effective label entropy} by reducing the number of equally plausible surface realizations per step, (ii) \emph{increases information density} at reasoning checkpoints via clearer demarcation and explicit unit/constraint checks, and (iii) \emph{reduces gradient noise} induced by heterogeneous narration. Together, these factors yield faster descent and a lower asymptote in cross-entropy, aligning with the test improvements on \textsc{AoPS2024}/\textsc{OlympiadBench} (Table~\ref{math_result}). While training loss alone does not guarantee generalization, the consistent pattern—lower loss, smoother convergence, and higher accuracy—supports DTA as a more learnable and higher-signal supervision scheme in the distillation-free setting.

\subsubsection{Investigating Latent Space Shift}
\begin{figure}[t]
	\centering
	\includegraphics[width=0.5\textwidth]{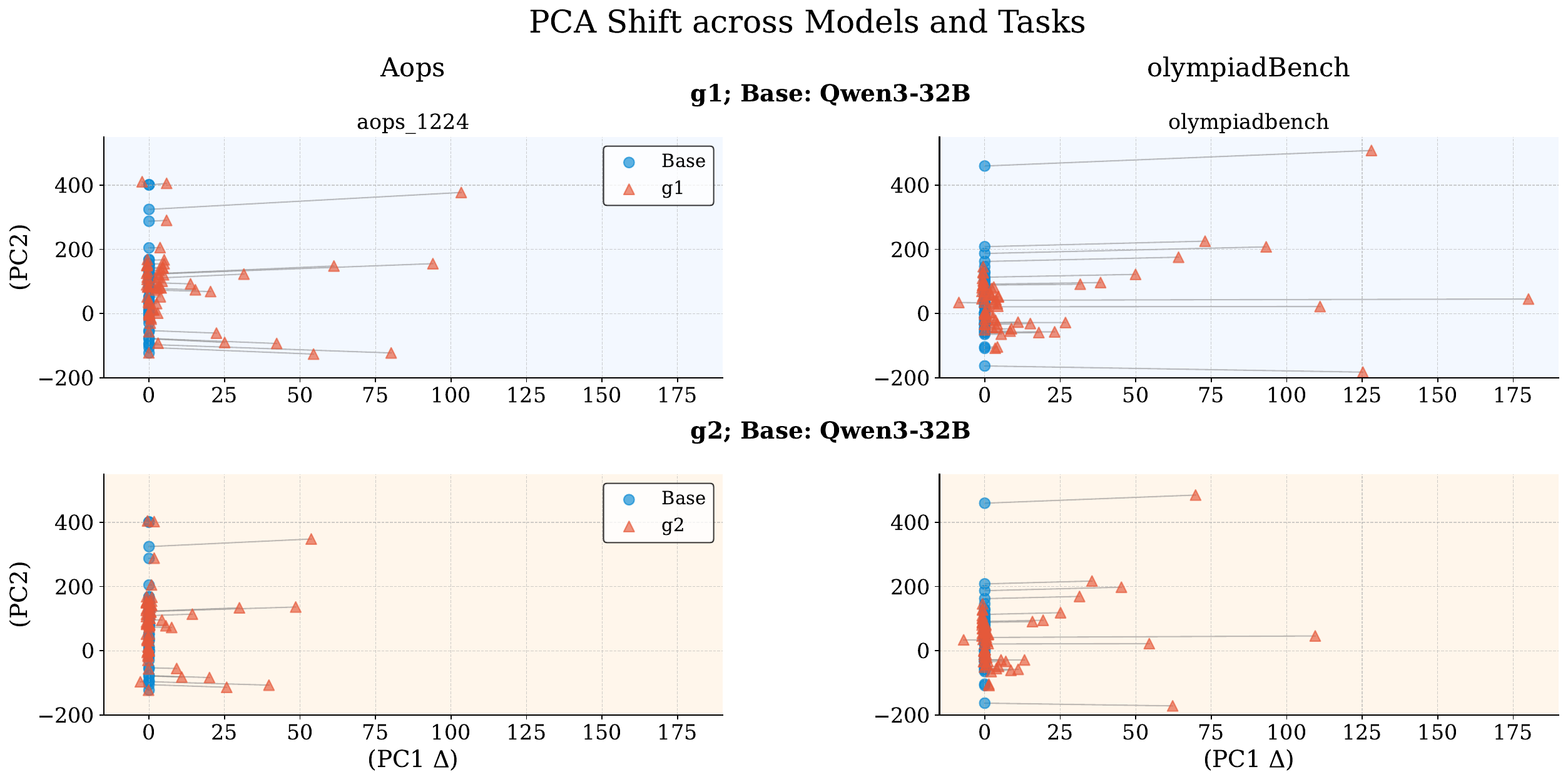}
	\caption{PCA shift of \textsc{Qwen3-32B} across training methods and benchmarks.}
	\label{math_PCA_shift}
\end{figure}
Motivated by recent analyses of representation drift~\cite{huan2025doesmathreasoningimprove}, we project hidden states onto leading principal components to visualize training-induced shifts (Figure~\ref{math_PCA_shift}). Qualitatively, $g2_{\text{gene}}$ shows \emph{smaller} displacement from the base model than $g1_{\text{gene}}$ on both datasets, indicating that DTA preserves the base model’s latent geometry more faithfully. We interpret this as evidence that (i) standard SFT on mixed-quality rationales can behave like a \emph{wholesale reconstruction} that perturbs broad linguistic/semantic subspaces, whereas (ii) DTA effects \emph{targeted refinement}—strengthening reasoning-relevant pathways while limiting collateral drift. This helps explain why $g2_{\text{gene}}$ improves accuracy without requiring explicit test-time chain-of-thought: underlying circuits leveraged by the base are preserved and sharpened rather than overwritten.

\subsubsection{Token Shift Analysis}

\begin{figure}[t]
	\centering
	\subfigure[$g1_{\text{gene}}$]{\includegraphics[width=0.2\textwidth]{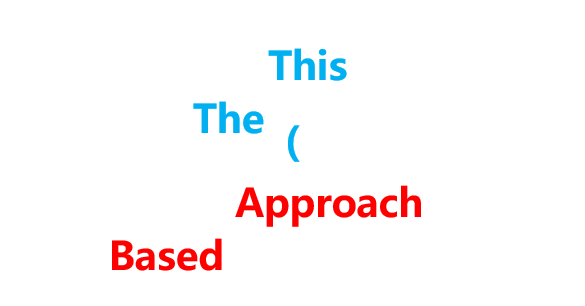}}
	\subfigure[$g2_{\text{gene}}$]{\includegraphics[width=0.2\textwidth]{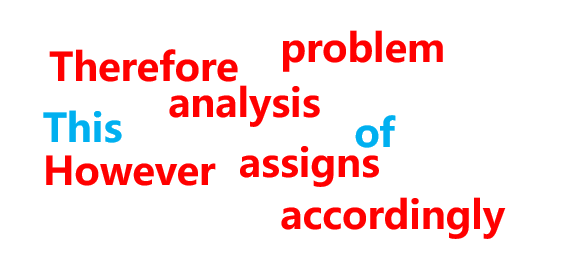}}
	\caption{Word clouds of token shift in the ablation.}
	\label{token shift ablation}	
\end{figure}
Figure~\ref{token shift ablation} highlights systematic differences in token shift. Relative to the base, $g2_{\text{gene}}$ increases the prevalence of \emph{logical–structural} discourse markers (e.g., \emph{therefore}, \emph{however}, \emph{hence}), whereas $g1_{\text{gene}}$ concentrates shifts on \emph{content} nouns/numerals. Emphasizing scaffolding tokens aligns with more explicit step demarcation and hypothesis–test–conclude patterns, which (i) reduce local ambiguities in multi-step algebra, (ii) stabilize constraint propagation (units, bounds, parity), and (iii) improve the verifiability of intermediate claims. 

\subsubsection{Ablation Summary}
Across both benchmarks, the convergence of evidence—accuracy gains, preference wins, smaller representation drift, and increased logical markers—supports the claim that trajectory alignment improves general mathematical reasoning by shaping \emph{how} solutions are narrated, not merely \emph{what} facts are recalled. These findings suggest that trajectory-aware curation is a complementary lever to scale and token count for improving generalization in math-heavy domains.

\section{Conclusion}
We introduced \emph{Data Trajectory Alignment} (DTA), a two-phase, model-agnostic curation framework that treats \emph{solution processes} as first-class supervision for domain adaptation of LLMs. By rewriting teacher trajectories to match the student’s inductive biases and selecting high-signal exemplars with reflection-based judging, DTA improves first-pass correctness without enabling explicit “thinking” modes. On \textsc{Telemath}, the DTA-trained model $g2_{\text{tele}}$ attains state-of-the-art accuracy (Table~\ref{tab:leaderboard}), surpassing the distilled-only variant ($g1_{\text{tele}}$) by +17.65 pass@1 and outperforming a strong thinking-enabled baseline by +2.94 pass@1—while preserving inference simplicity. Token-shift analysis (Figure~\ref{token shift telemath}) shows that DTA increases logical–structural markers rather than merely amplifying domain nouns, and representation analyses in the distillation-free ablation suggest targeted refinement rather than wholesale rewriting, aligning with the observed accuracy and robustness gains (Table~\ref{math_result}, Figure~\ref{math_PCA_shift}).

From a systems perspective, DTA moves cost from \emph{inference-time sampling} to \emph{training-time curation}. Under edge-like settings (Sec.~\ref{inference_setting}), $g2_{\text{tele}}$ reduces energy per token, latency, and EDP simultaneously (Table~\ref{mobile}, Table~\ref{tab:mobile_derived}), yielding \emph{2--4$\times$} lower time/energy per correct answer than strong non-thinking baselines, and \emph{16--20$\times$} than thinking-mode decoding. These efficiency gains arise from higher pass@1 (less re-sampling), stable step demarcation (fewer meanders), and higher steady-state decode rates—properties that are directly valuable for mobile/edge deployments.

Overall, our results indicate that \emph{how} supervision is curated—at the trajectory level—can substitute for expensive inference-time reasoning in constrained environments. DTA offers a practical recipe for low-resource verticals: align intermediate steps and presentation style to the target student, filter by informativeness and reward, and harvest compact, high-yield training signals that are both accuracy- and efficiency-effective.

\section{Limitations and Future Work}
\textbf{Measurement and evaluation.} (i) Energy is measured on the GPU only via \texttt{nvidia-smi dmon} at 10\,Hz; host/DRAM/network power and short bursts may be under-captured in our setup. (ii) Results are on A800 GPUs with fixed runtime (\texttt{vLLM==0.9.1}), batch size, and no dynamic batching; absolute numbers may vary with hardware, DVFS, and server settings. (iii) Derived metrics (e.g., EDP proxy, time/energy per correct) assume comparable output structure across non-thinking systems; although we report tokens/sample for context, residual length effects can remain.

\textbf{Methodological constraints.} (i) Quality gates rely on LLM judges and a reversed-IFD signal; correlated teacher/judge errors or rubric bias can leak through. (ii) The reward aggregation weights are fixed and may not be optimal across tasks. (iii) DTA aligns to a single student style; over-regularization of discourse markers could reduce stylistic diversity or harm transfer in tasks that benefit from alternative reasoning forms.

\textbf{Data scope and external validity.} (i) Experiments focus on English and two domains (telecom mathematics and general math); broader multilingual, multimodal (tables/diagrams), or regulation-heavy settings are not yet covered. (ii) Despite MinHash-based decontamination, residual near-duplicates cannot be fully excluded. (iii) Edge results reflect a server-side approximation of mobile constraints rather than true on-device measurements.

\textbf{Future work.} We see three co-design directions. (1) \emph{Verification-aware DTA:} integrate formal/unit checkers, executable math, and programmatic tests into the reward to reduce judge bias and improve faithfulness. (2) \emph{Hardware-/efficiency-aware selection:} fold latency/energy (e.g., EDP proxy) into candidate ranking; explore difficulty-conditioned style alignment, output-budget shaping, and early-exit–friendly formats that further reduce tokens/sample. (3) \emph{Deployment realism:} extend to quantization-aware DTA, heterogeneous hardware (mobile NPUs), streaming/online decoding, multilingual and multimodal trajectories, and safety/calibration audits (uncertainty, abstention). We also plan comprehensive ablations on batch sizing, dynamic batching, and decoding policies, plus human evaluation for long-horizon tasks.

In summary, while DTA already closes the accuracy–efficiency gap without relying on thinking-mode decoding, scaling it with verification signals and hardware-in-the-loop objectives is a promising path toward dependable, resource-efficient foundation models for mobile/edge applications.

\bibliographystyle{IEEEtran}
\bibliography{ref}

\appendix
\subsection{Detailed Solution Generation Prompt Template}
\label{Detail Prompt Template}
\begin{framed}
	\setlength{\parindent}{0pt}
	 You are an analysis expert. Based on the \textit{[domain]} problem provided by the user and the correct answer to the questions, you need to rewrite the answer into detailed solution that include analytical content and output them.
	\medskip
	\textbf{Input Information}
	\begin{itemize}
		\item The full question
		\item The correct answer
	\end{itemize}
	
	\textbf{Rules}
	\begin{itemize}
		\item The rewritten correct answer must include rigorous logical reasoning.
		\item The rewritten paragraph must strictly adhere to the correct answer of the input question.
		\item Write down the analysis process step by step.
	\end{itemize}
	
	\textbf{The Full Question}
	
	\textit{[the question provided]}
	
	\textbf{The Correct Answer}
	
	\textit{[the correct answer provided]}
\end{framed}
\noindent where \textit{domain} is the domain of input sample, \textit{the question provided} and \textit{the correct answer provided} are $(x_{seed}, y_{seed})$.

\subsection{Knowledge \& Skill Extraction and Summary Prompt Template}
\label{Extract Prompt Template}
\begin{framed}
	\setlength{\parindent}{0pt}
	You are an expert in \textit{[domain]}. Your task is to analyze provided problems and their detailed solutions to extract and summarize two key components: the underlying knowledge points and the problem-solving skills used.

	\textbf{Input Information}
	\begin{itemize}
		\item The full question
		\item The detailed solution
	\end{itemize}
	
	\textbf{Core Knowledge Points Assessed}
	\begin{itemize}
		\item Identify and list the fundamental concepts, theories from the field of \textit{[domain]} that the question is designed to test.
		\item Be specific about the concepts and theories
		\item Explain why each knowledge point is relevant to the problem.
	\end{itemize}
	
	\textbf{Problem-Solving Skills}
	\begin{itemize}
		\item Describe the high-level strategy used to tackle the problem.
		\item How was the complex problem decomposed into smaller, manageable parts?
		\item List all key formulas, equations, and mathematical principles applied in the solution.
		\item For each formula, state its purpose in the context of the problem
		\item Mention any specific constants or standard values used
	\end{itemize}
	
	\textbf{Output Format}
	\begin{itemize}
		\item Core Knowledge Points Assessed
		\begin{itemize}
			\item \textbf{Knowledge Point 1:} [Description]. Relevance: [Explanation].
			\item \textbf{Knowledge Point 2:} [Description]. Relevance: [Explanation].
			\item ... (add more as needed)
		\end{itemize}
		\item Problem-Solving Skills
		\begin{itemize}
			\item \textbf{Strategy:} [Description]
			\item \textbf{Decomposing:} [Description]. Relevance: [Explanation].
			\item \textbf{Formula Application \& Mathematical Tools:} [Description].
		\end{itemize}
	\end{itemize}
	
	\textbf{The Full Question}
	
	\textit{[the question provided]}
	
	\textbf{The Correct Answer}
	
	\textit{[the correct answer provided]}
\end{framed} 
\noindent where \textit{domain} is the domain of input sample, \textit{the question provided} and \textit{the correct answer provided} are $(x_{seed}, y_{seed_{i}}^{*})$.

\subsection{Training Data Generation Prompt Template}
\label{Generate Prompt Template}
\begin{framed}
	\setlength{\parindent}{0pt}
	\noindent You are an expert educational content creator with a deep specialization in \textit{[domain]}. Your task is to generate unique, challenging, and pedagogically sound problems based on specified knowledge points and a set of problem-solving skills.
	
	\textbf{Input Information}
	\begin{itemize}
		\item The knowledge points: [point 1], [point 2], [point 3] ...
		\item The problem-solving skills: [skill 1], [skill 2] ...
	\end{itemize}
	
	\textbf{Rules}
	\begin{itemize}
		\item The problem must be a real-world scenario from the given knowledge points, incorporating all technique from the problem-solving skills, and uses practical values from \textit{[domain]}.
		\item The problem statement must be unambiguous. All variables and constants required for the solution must be provided within the problem statement.
		\item All technical details, formulas, and standard values must be used correctly.
		\item Vary the parameters to create a unique problem each time.
	\end{itemize}
	
	\textbf{Generate the output in the following exact format}
	\begin{itemize}
		\item Problem Statement: [A clear, self-contained description of the scenario and the question to be solved. Include all necessary data with units. The problem should be solvable using the provided data and require a mathematical calculation.]
		\item Solution Steps: [A step-by-step, numbered list explaining the reasoning and mathematical operations required to solve the problem. Do not just show the calculation; explain the "why" behind each step.]
		\item Final Answer: [The final numerical answer, must be a single numerical value (integer or float)]
	\end{itemize}
	
	\textbf{The Knowledge Point}
	
	\textit{[the knowledge points]}
	
	\textbf{The Problem-Solving Skills}
	
	\textit{[the problem-solving skills]}
\end{framed}

\subsection{Peer Review Prompt Template}
\label{peer review prompt}
\begin{framed}
	\setlength{\parindent}{0pt}
	\noindent You are an impartial grader. Your task is to evaluate whether the Answer correctly and completely addresses the Question. Focus \textbf{only} on factual/logic correctness and task completion—not style.
	
	\textbf{Evaluation Rules}
	\begin{itemize}
		\item Verify factual claims using generally accepted knowledge and deductive reasoning. Recompute any math step-by-step; check units and boundary conditions.
		\item For multi-part questions, require each requested item; partial credit is allowed.
		\item Penalize contradictions, hallucinations, unjustified assumptions, or failure to follow constraints in the Question.
		\item If the Question is inherently subjective/opinion-based or lacks enough information to determine correctness, treat it as not objectively verifiable and return 0.00.
	\end{itemize}
	
	\textbf{Scoring Rubric (Continuous)}
	\begin{itemize}
		\item 1.00: Fully correct and complete; no incorrect claims.
		\item 0.75: Mostly correct; at most minor omission/benign slip.
		\item 0.50: Partly correct; significant omission or one major error.
		\item 0.25: Small fragments correct; mostly incorrect/irrelevant.
		\item 0.00: Incorrect, off-topic, or not objectively verifiable.
	\end{itemize}
	
	\textbf{Output format}
	
	First output the reasoning process then the score in \textbackslash box{}, rounded to two decimals (e.g., \textbackslash box{0.83}).
	
	\textbf{Input Question}
	[Question]
	
	\textbf{Input Answer}
	[Answer]
\end{framed} 

\subsection{Summarize Model Answer Style Prompt Template} 
\label{Summarize Prompt Template}
\begin{framed}
	\setlength{\parindent}{0pt}
	You are a style inducer for domain answers, you will be given a set of domain-specific Q \& A pairs that exemplify how answers should look in this domain.
	
	\textbf{TASK}
	\begin{itemize}
		\item Infer and summarize the answering style from the examples.
		\item Convert it into a self-contained, prescriptive style guide that can be used as a prompt for future answers in this domain.
	\end{itemize}
	
	\textbf{CRITICAL REQUIREMENTS}
	\begin{itemize}
		\item Evidence-only policy: Every rule you output must be directly induced from patterns present in the provided Q \& A pairs. Do NOT invent rules, add “best practices,” or import outside knowledge. If a requirement is not evidenced by the examples, omit it.
		\item Write the output as imperative rules (e.g., “Use…”, “Avoid…”, “Limit…”). Do NOT describe the examples; prescribe behavior.
		\item Be concrete and testable (e.g., “1–2 sentence summary up front”, “bulleted lists for at least 3 items”, “define jargon on first use”).
		\item Do not reference “the examples above,” “training data,” or “as shown.” The guide must stand alone.
		\item Resolve inconsistencies by preferring the most frequent patterns across the examples; if no clear majority exists, omit the conflicting rule.
		\item Include low-frequency patterns only if they clearly appear in at least one example and improve clarity or safety in the same domain.
	\end{itemize}
	
	\textbf{OUTPUT FORMAT} (Markdown). 
	
	Produce ONLY these two sections:
	\begin{itemize}
		\item Language, Tone, and Level of Detail — Requirements
		\begin{itemize}
			\item Bullet rules about language/register (e.g., formal vs. informal, first/second person, jargon usage and definitions, hedging vs. confidence, citation habits, disclaimers/safety notes if applicable).
			\item Bullet rules about tone (e.g., neutral/objective, empathetic, authoritative, motivational).
			\item Bullet rules about level of detail (e.g., step-by-step vs. high-level, when to add examples, equations, code, or edge cases; typical answer length ranges; when to provide references).
			\item Explicit “Avoid” items (e.g., speculation, unsupported claims, slang, verbosity, fragile assumptions).
		\end{itemize}
		\item Answer Structure and Organization — Requirements
		\begin{itemize}
			\item Opening pattern: e.g., one-sentence answer or TL;DR summary.
			\item Core explanation: ordering principles (from basics→advanced, most-common→edge cases, or decision tree).
			\item Formatting rules: headings usage, numbered steps vs. bullets, tables when comparing no less than 2 options, inline definitions for key terms, callouts for warnings/assumptions.
			\item Evidence \& attribution: when/how to cite, link, or name sources.
			\item Actionability: checklists, examples, templates; when to include them.
			\item Closers: brief recap, next steps, caveats, or pointers to further reading.
			\item Adaptation rules: how to shorten for simple questions; how to deepen for expert queries.
		\end{itemize}
	\end{itemize}
	
	\textbf{INFERENCE PROCEDURE} (follow quietly; do not output this section)
	\begin{itemize}
		\item Scan all answers to tally recurring linguistic and structural patterns.
		\item Only consider a candidate rule if it is directly observable in at least one example; prefer rules supported by multiple examples.
		\item When patterns conflict, favor the majority pattern; if ties persist, omit the rule rather than guessing.
		\item Generalize only within the domain reflected by the examples; do not extrapolate beyond observed behaviors.
	\end{itemize}
	
	\textbf{DELIVERABLE}
	Return only the two Markdown sections above. No preamble, no conclusion.
	
	\textbf{EXAMPLES BLOCK}
	[Insert domain Q \& A pairs here]
\end{framed}

\subsection{Construct Trajectory Aligned Data Prompt Template}
\label{Align Prompt Template}
\begin{framed}
	\setlength{\parindent}{0pt}
	\noindent You are an analysis expert. Based on the problem provided and the corresponding correct answer, you need to rewrite the solution in details following the rules and format requirements.
	
	\textbf{Input Information}
	\begin{itemize}
		\item The full question
		\item The correct answer
	\end{itemize}
	
	\textbf{Rules}
	\begin{itemize}
		\item The rewritten solution must strictly adhere to the correct answer of the input question.
		\item \textit{[additional rules related to language, tone \& level of details]}
	\end{itemize}
	
	\textbf{Format of the Detailed Answer}
	
	\textit{[requirements about the answer structure \& organization]}
	
	\textbf{The Full Question}
	
	\textit{[the question provided]}
	
	\textbf{The Correct Answer}
	
	\textit{[the correct answer provided]}
\end{framed}
\noindent The \textit{additional rules related to language, tone \& level of details} and \textit{requirements about the answer structure \& organization} are the information of answer style $S_{a}$, \textit{the question provided} and \textit{the correct answer provided} are original training sample $(x_{train}, y_{train})$.

\subsection{Reflection on Reward Model Prompt Template}
\label{reflection prompt}
\begin{framed}
	\setlength{\parindent}{0pt}
	You are an impartial grader. Evaluate the \textbf{Answer} to the \textbf{Question} strictly by the rubric below. Do not solve the task yourself beyond what is necessary to assess quality. Be conservative and evidence-based.
	
	\medskip
	\textbf{Inputs}
	\begin{itemize}
		\item \textbf{Question}: \texttt{\{\{QUESTION\}\}}
		\item \textbf{Answer}: \texttt{\{\{ANSWER\}\}}
	\end{itemize}
	
	\textbf{Rubric (0--10 each)}
	
	\medskip
	\textbf{1) Correctness (facts \& hallucinations)}
	\begin{itemize}
		\item 10: All claims are factually correct; no contradictions; no hallucinations.
		\item 8: Minor, non-critical imprecision; core claims are correct.
		\item 5: Mixed correctness; at least one significant error or unsupported claim.
		\item 2: Mostly incorrect or speculative; multiple unsupported claims.
		\item 0: Entirely incorrect, fabricated, or non-responsive.
	\end{itemize}
	
	\noindent\textbf{2) Completeness (coverage of the query)}
	\begin{itemize}
		\item 10: Addresses every part of the Question; no missing steps/results; includes necessary caveats.
		\item 8: Covers most parts; minor omissions that don’t affect the main outcome.
		\item 5: Partial coverage; important subparts or edge cases are missing.
		\item 2: Barely addresses the request; major gaps.
		\item 0: Does not address the request at all.
	\end{itemize}
	
	\noindent\textbf{3) Clarity (coherence \& organization)}
	\begin{itemize}
		\item 10: Well-structured, easy to follow; precise wording; unambiguous terms.
		\item 8: Generally clear with minor awkwardness or minor organizational issues.
		\item 5: Understandable but disorganized, verbose, or occasionally confusing.
		\item 2: Hard to follow; poor structure; ambiguous phrasing.
		\item 0: Incoherent.
	\end{itemize}
	
	\noindent\textbf{4) Conciseness (no redundancy)}
	\begin{itemize}
		\item 10: Only necessary information; no fluff or repetition.
		\item 8: Minor redundancy; mostly succinct.
		\item 5: Noticeable verbosity or repetition.
		\item 2: Highly verbose; much irrelevant content.
		\item 0: Rambling or padded with irrelevant text.
	\end{itemize}
	
	\textbf{Scoring Guidance}
	\begin{itemize}
		\item Use whole numbers 0--10 for each category.
		\item Base judgments on the \textbf{Answer} with respect to the \textbf{Question}. If factual verification is uncertain, penalize Correctness proportionally to the risk of hallucination.
		\item The \textbf{Overall} score is the unweighted average of the four category scores, rounded to one decimal.
	\end{itemize}
	
	\textbf{Output Format (JSON only)}\\
	Return \textbf{only} a JSON object with this schema:
	\begin{verbatim}
		{
			"correctness": { 
				"score": 0-10, 
				"explanation": "1 or 2 sentence 
				justification" 
				},
			"completeness": { 
				"score": 0-10, 
				"explanation": "1 or 2 sentence 
				justification" 
				},
			"clarity": { 
				"score": 0-10, 
				"explanation": "1 or 2 sentence 
				justification" 
				},
			"conciseness": { 
				"score": 0-10, 
				"explanation": "1 or 2 sentence 
				justification" 
				},
		}
	\end{verbatim}
	\textbf{Constraints}
	\begin{itemize}
		\item Keep explanations brief (no step-by-step reasoning or internal chain-of-thought).
		\item Do not include any text outside the JSON.
	\end{itemize}
\end{framed}

\subsection{The Pair-Wise Comparison}
\label{pairwise}
The prompt template is shown below:
\begin{framed}
	\setlength{\parindent}{0pt}
	\textbf{Role \& Task:}
	
	You are an impartial and highly accurate evaluation model. Your task is to critically assess two provided answers (Answer A and Answer B) to a given question. You must assign a score from 0 to 10 to each answer, where 0 is the worst and 10 is the best.
	Please provide a comprehensive explanation of your evaluation, avoiding any potential bias and ensuring that the order in which the responses were presented does not affect your judgment.
	The scores must be justified based on the following criteria.
	
	\textbf{Evaluation Criteria:}
	\begin{itemize}
		\item Helpfulness: Does the answer effectively address the user's implied need behind the question? Is it useful, actionable, and does it solve the user's potential problem?
		\item Relevance: Does the answer stay directly on-topic? Does it address the question asked without introducing significant irrelevant information or digressions?
		\item Correctness \& Factual Accuracy: Is the information provided factually sound, verifiable, and logically consistent? Are any assumptions or uncertainties clearly stated? This is the most critical criterion; severe factual errors must result in a low score.
		\item Level of Detail \& Completeness: Does the answer provide sufficient depth and breadth? Does it cover the main aspects of the question without being overly verbose or unnecessarily sparse? A good answer is thorough yet concise.
		\item Clarity \& Structure: Is the answer well-organized and easy to understand? Is it presented in a logical manner (e.g., using paragraphs, bullet points, or step-by-step explanations)?
	\end{itemize}
	
	\textbf{Scoring Guide:}
	\begin{itemize}
		\item 0-3 (Poor): Incorrect, irrelevant, unhelpful, or nonsensical.
		\item 4-5 (Adequate): Partially addresses the question but has major flaws: significant inaccuracies, lack of detail, or poor structure.
		\item 6-7 (Good): Correct and relevant, but may be incomplete, lack depth, or have minor inaccuracies. It is helpful but not comprehensive.
		\item 8-9 (Very Good): Accurate, relevant, detailed, and well-structured. It is clearly helpful and would satisfy most users.
		\item 10 (Excellent): Exceptional in all criteria. Perfectly accurate, highly comprehensive, impeccably structured, and anticipates potential follow-up questions. It provides exceptional value.
	\end{itemize}
	
	\textbf{Output Format:}
	You must provide your evaluation in the following exact JSON structure. Do not include any other text before or after the JSON object.
	\begin{verbatim}
	{
		"evaluation": {
			"answer_a": {
				"score": <number>,
				"justification": "<concise paragraph 
				explaining the score based on the 
				criteria>"
			},
			"answer_b": {
				"score": <number>,
				"justification": "<concise paragraph 
				explaining the score based 
				on the criteria>"
			},
			"summary": "<A brief overall summary
			comparing both answers and stating
			which one is superior,
			if applicable.>"
		}
	}
	\end{verbatim}
	
	\textbf{Process:}
	\begin{itemize}
		\item Analyze the user's question carefully.
		\item Evaluate Answer A rigorously against all five criteria.
		\item Assign a score to Answer A and draft a justification.
		\item Evaluate Answer B rigorously against all five criteria, without being biased by your assessment of Answer A.
		\item Assign a score to Answer B and draft a justification.
		\item Write a final summary comparing the two.
		\item Format your response strictly as JSON.
	\end{itemize}
	
	\textbf{Input:}
	\begin{itemize}
		\item Question: \
		[The user's question will be inserted here]
		\item Answer A: \
		[The first answer to evaluate]
		\item Answer B: \
		[The second answer to evaluate]
	\end{itemize}
	
	Now, begin your evaluation for the provided Question, Answer A, and Answer B.
\end{framed}

To further mitigate the positional bias elaborated upon in \cite{wang2023largelanguagemodelsfair}, we follow the implementation from Cherry LLM \cite{li2024quantityqualityboostingllm} by reversing Answer A and Answer B in prompt above, which greatly eliminates the potential position bias of GPT5.
\end{document}